\documentclass[runningheads]{llncs}
\usepackage{graphicx}
\usepackage[misc]{ifsym}
\usepackage{graphicx, amsmath, amsfonts, amssymb, xspace, color, booktabs, hyperref, setspace, subcaption}

\usepackage{amsthm}
\usepackage{xcolor}
\usepackage[noend,ruled,linesnumbered]{algorithm2e}
\usepackage{multirow}

\newcommand{\ie}{\emph{i.e.}\xspace} 
\newcommand{\eg}{\emph{e.g.}\xspace} 
\newcommand{\our}{\textsc{FUSE}\xspace}
\newcommand{\nop}[1]{}

\begin{document}
\title{FUSE: Multi-Faceted Set Expansion by Coherent Clustering of Skip-grams\thanks{Accepted at ECML-PKDD 2020}}
%
\author{Wanzheng Zhu\inst{1}  [\Letter] \and
Hongyu Gong\inst{1} \and
Jiaming Shen\inst{1} \and
Chao Zhang\inst{2} \and
Jingbo Shang\inst{3} \and
Suma Bhat\inst{1} \and
Jiawei Han\inst{1}}
\authorrunning{W. Zhu et al.}
%
\institute{University of Illinois at Urbana-Champaign, IL, USA \\
	\email{\{wz6, hgong6, js2, spbhat2, hanj\}@illinois.edu}\\ \and
	Georgia Institute of Technology, GA, USA \\
	\email{chaozhang@gatech.edu}\\ \and
	University of California San Diego, CA, USA\\
	\email{jshang@ucsd.edu}}
\maketitle              
\begin{abstract}
	Set expansion aims to expand a small set of seed entities into a complete set of relevant entities. 
	Most existing approaches assume the input seed set is unambiguous and completely ignore the multi-faceted semantics of seed entities.
	As a result, given the seed set \{``Canon'', ``Sony'', ``Nikon''\}, previous models return \emph{one mixed set} of entities that are either \textit{Camera Brands} or \textit{Japanese Companies}. 
	In this paper, we study the task of \textbf{multi-faceted set expansion}, which aims to capture all semantic facets in the seed set and return \emph{multiple sets} of entities, one for each semantic facet. 
	We propose an unsupervised framework, \our, which consists of three major components: 
	(1) \emph{facet discovery module}: identifies all semantic facets of each seed entity by extracting and clustering its skip-grams, and 
	(2) \emph{facet fusion module}: discovers shared semantic facets of the entire seed set by an optimization formulation, and
	(3) \emph{entity expansion module}: expands each semantic facet by utilizing a masked language model with pre-trained BERT models. 
	Extensive experiments demonstrate that \our can accurately identify multiple semantic facets of the seed set and generate quality entities for each facet. 
	
	\vspace{-0.1cm}
\end{abstract}

\section{Introduction}
The task of \textit{set expansion} is to expand a small set of seed entities into a more complete set of relevant entities. 
For example, to explore all \textit{Universities in the U.S.}, one can feed a seed set (\eg, \{``Stanford", ``UCB'', ``Harvard''\}) to a set expansion system and then expect outputs such as ``Princeton'', ``MIT" and ``UW''. 
Those expanded entities can benefit numerous entity-aware applications, including query suggestion \cite{cao2008context}, taxonomy construction \cite{velardi2013ontolearn}, recommendation \cite{zhu2018spherical}, and information extraction \cite{jain2010open,sarker2015portable,weikum2010information,zhao2014knowledge}. Besides, the set expansion algorithm itself becomes a basic building block of many NLP-based systems \cite{Mamou2018TermSE,Shen2018HiExpanTT}. 

Previous studies on set expansion focus on returning \emph{one single set} of most relevant entities. 
Methods have been developed to incrementally and iteratively add the entities of high confidence scores into the set. 
A variety of features are extracted, including word co-occurrence statistics \cite{Pantel2009WebScaleDS}, unary patterns \cite{Shen2017SetExpanCS}, or coordinational patterns \cite{Sarmento2007MoreLT}, from different data sources such as query log \cite{tong2008system}, web table \cite{Wang2015ConceptEU}, and raw text corpus \cite{Mamou2018TermSE,Shen2017SetExpanCS}. 
However, all these methods assume the given seed set is unambiguous and completely ignore the multi-faceted semantics of seed entities. 
As a result, given a seed set \{``apollo", ``artemis", ``poseidon"\} which has two semantic facets -- \textit{Major Gods in Greek Mythology} and \textit{NASA Missions}, previous methods can only generate one mixed set of entities from these two facets, which inevitably hampers their applicabilities.

\begin{figure}[th]
	\centering
	\includegraphics[width=0.7\linewidth]{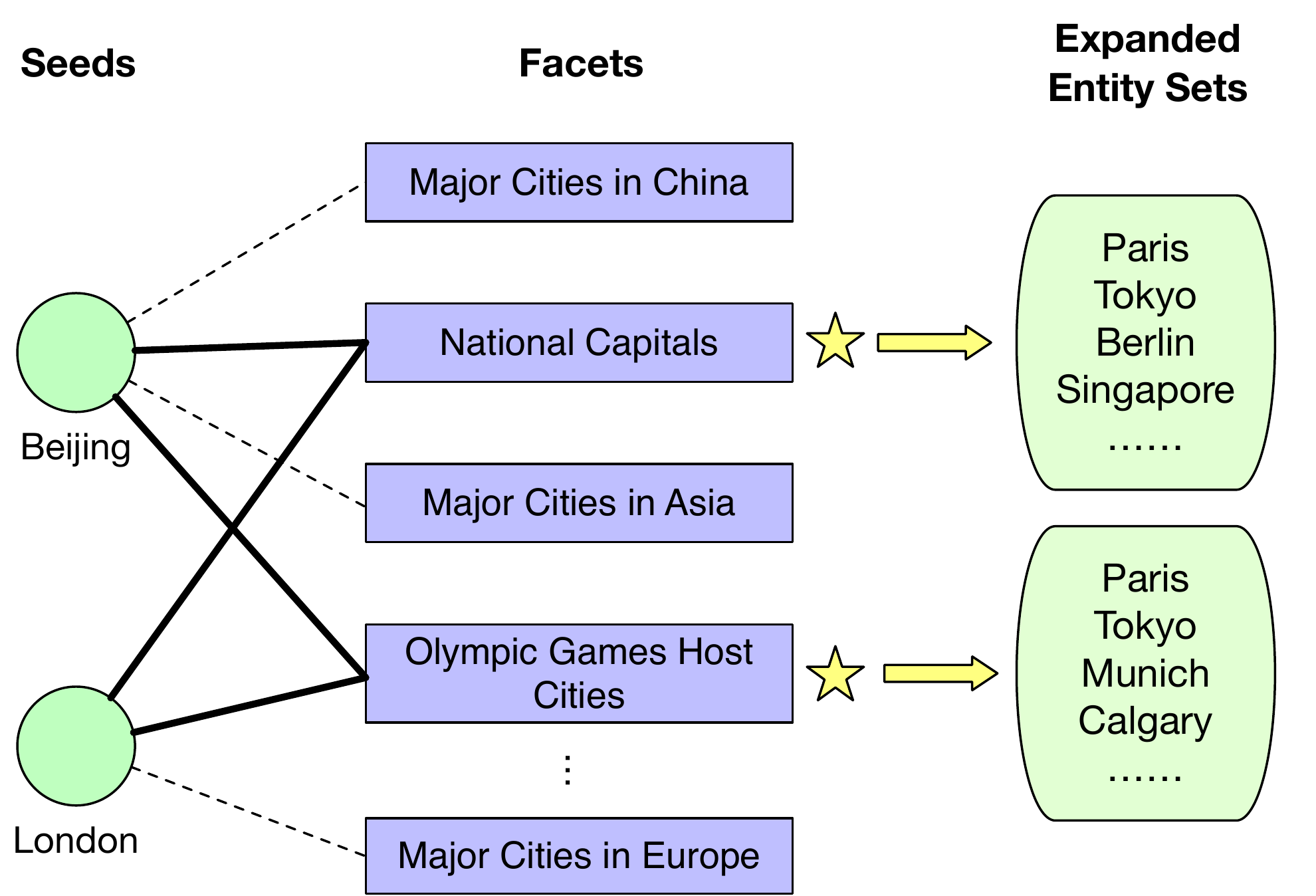}
	\caption{An illustrative example of a multi-faceted seed set \{``Beijing", ``London"\}. 
		Facets (\eg \textit{Major Cities in China}) that do not appear in both seed entities should be eliminated in the set expansion process.
		As a result, we expect to output two separate entity sets: one for semantic facet \textit{National Capitals} and the other one for semantic facet \textit{Olympic Games Host Cities}.}
	\label{fig:1}
\end{figure}

\begin{figure*}[t]
	\centering
	\includegraphics[width=0.99\linewidth]{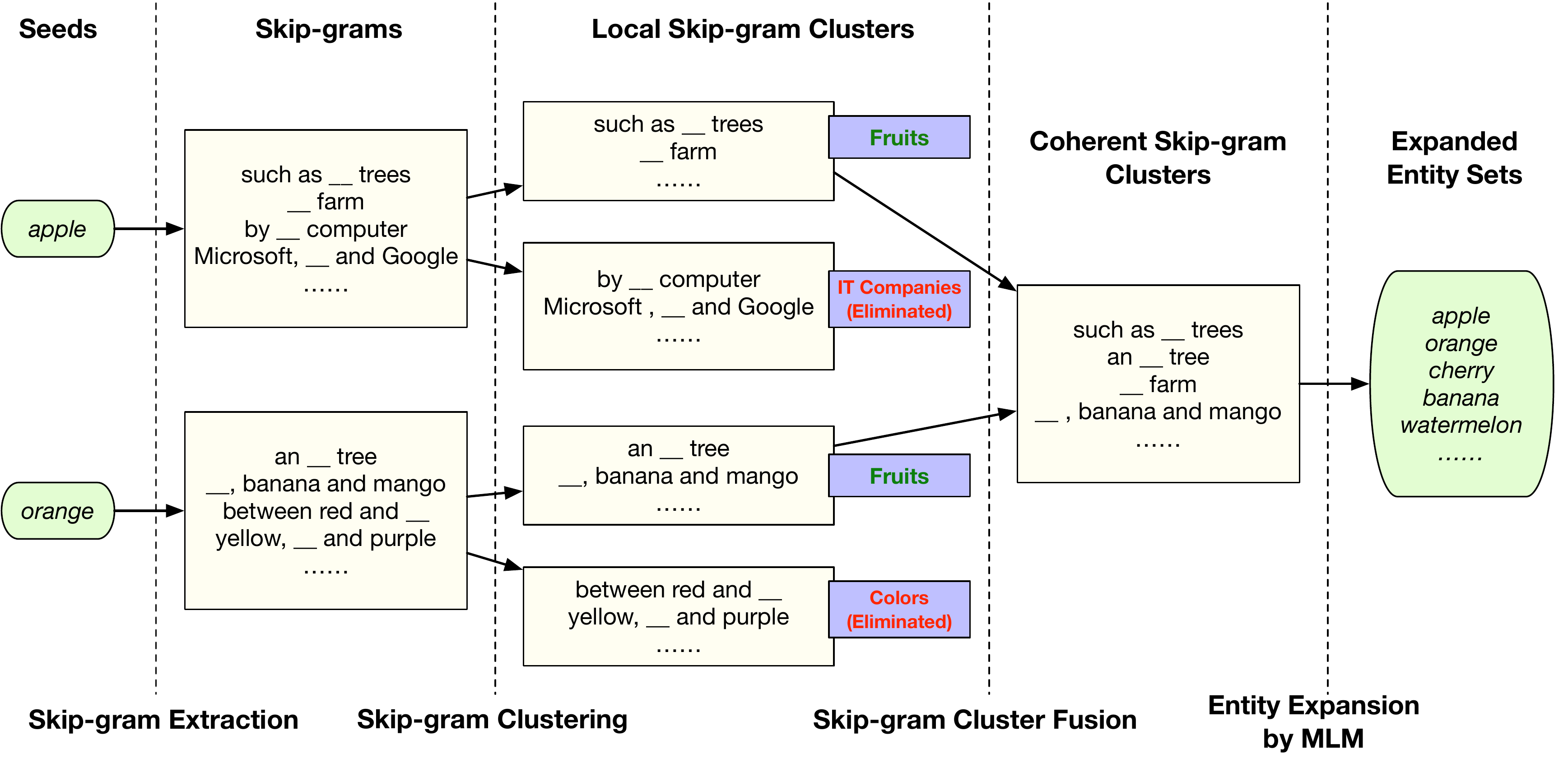}
	\caption{Overview of \our. The key novelty is to discover coherent skip-gram clusters, whereas previous methods skip this stage and directly combine all skip-grams into the same pool for entity expansion.}
	\label{fig:2}
\end{figure*}

In this paper, we approach the set expansion task from a new angle. 
Our study focuses on \textbf{multi-faceted set expansion} which aims to identify semantic facets shared by all seed entities and return \textit{multiple} expanded sets, one for each semantic facet.
The key challenge lies in the discovery of shared semantic facets from a seed set. 
However, the only initial attempt towards multi-facetedness, EgoSet \cite{rong2016egoset}, 
not only requires user-created ontologies as external knowledge, 
but also has no guarantee that their generated semantic facets are relevant to all seed entities.
As an illustrative example in Fig.~\ref{fig:1}, EgoSet generates more than five facets, but only two of them are relevant to both seeds.

To handle the key challenge of multi-faceted set expansion, we propose a novel framework, \our, as illustrated in Fig.~\ref{fig:2}.
First, we discover all possible facets of each seed by extracting and clustering its skip-grams.
Second, we leverage an optimization formulation to discover the shared semantic facets across all seeds as \textit{coherent semantic facets}. 
This helps eliminate those facets relevant only to a partial set of seeds. 
Third, based on the coherent skip-gram clusters, we utilize a masked language model (MLM) with pre-trained BERT models to generate quality entities for each semantic facet.

It is considerably complicated to evaluate such multi-faceted set expansion task, mainly because we have no prior knowledge about the number of facets in a seed set. Therefore, we are likely to observe different number of facets between the generated result and the ground truth (\eg, the ground truth may have 3 facets, while the generated result has 4 facets.). 
Previously proposed metric Mean-MAP (MMAP) in \cite{rong2016egoset} only measures how many entities and facets in the ground truth are covered by the generated result. 
However, it fails to measure how noisy those generated facets are and thus it biases toward methods that output as many facets as possible. 
To overcome the intrinsic limitation of MMAP, we propose a more comprehensive evaluation metric, Best-Matching Average Precision (BMAP), that can not only capture the purity of generated facets but also their coverage of ground truth facets.

Our contributions are highlighted as follows.
\begin{itemize}
	\item We identify the key challenge of \textit{multi-faceted set expansion} and develop \our to address it.\footnote{The code is available at \url{https://github.com/WanzhengZhu/FUSE}.}
	\item We propose to determine semantic facets by clustering skip-gram contexts, and utilize an optimization formulation to discover coherent semantic facets. 
	\item We propose a novel evaluation metric for multi-faceted set expansion problem, which is shown to be a more comprehensive measure. 
	\item Extensive experiments demonstrate that our proposed framework outperforms state-of-the-art set expansion algorithms significantly. 
\end{itemize}

\section{Problem Formulation}
\label{sec:problem_formulation}
A facet refers to one semantic aspect or sense of seed words. 
For example, \emph{Fruit} and \emph{Technology Companies} are two facets of the word ``apple''. 
Previous works study mostly single-faceted set expansion and ignore the seeds' multi-facetedness nature.
In this work, we explore a better coverage of all coherent semantic facets of a seed set and study \textit{corpus-based multi-faceted set expansion}. 

More formally, given a seed set query $q=\{s_1, s_2, ..., s_m\}$ where $s_i$ is a seed and a raw text corpus $D$, our set expansion system is to find all lists of entities $\mathbb{E} = \{E^{(i)}, E^{(j)}, E^{(k)}, \dots \}$, where $E^{(i)}=\{x_{1}^{(i)}, \ldots, x_{n}^{(i)}\}$ is relevant to the $i$-th facet $f_i$ of query $q$, and $x_l^{(i)}$ denotes an expanded entity.

\section{Model}
\label{sec:model}

Our proposed \our framework consists of three main steps: 
1) extracting and clustering skip-grams for each seed (c.f. Sec~\ref{model_clustering});
2) discovering coherent semantic facets of a seed set (c.f. Sec~\ref{model_merging}); and
3) expanding entities for each semantic facet (c.f. Sec~\ref{model_expansion}).
An overview of our approach is shown in Fig.~\ref{fig:2} and the algorithm is shown in Algorithm \ref{algo}. 

\begin{algorithm}[!t]
	\caption{ \our: Multi-faceted Set Expansion}
	\label{algo}
	\KwIn{Corpus $D$; a user query $q$.}
	\KwOut{a list of expanded entity lists $\mathbb{E}$.}
	
	$\boxdot$ \emph{\textbf{Skip-gram Clustering}}\;
	seedClusterDict = \{\}\;
	\For{$seed$ in $q$} {
		sgs $\gets$ extractSkipgrams($seed$, $D$)\;
		sgClusters $\gets$ clustering(sgs)\;
		seedClusterDict[$seed$] $\gets$ sgClusters\;
	}
	
	$\boxdot$ \emph{\textbf{Clusters Fusion}}\;
	refSeed $\gets$ $q$.pop()\;
	refC $\gets$ seedClusterDict[refSeed]\;
	\While{$q$ is not empty} {
		curSeed $\gets$ $q$.pop()\;
		curC $\gets$ seedClusterDict[curSeed]\;
		coherentC $\gets$ fuseClusters(refC, curC)\;
		refC $\gets$ coherentC\;
	}
	
	$\boxdot$ \emph{\textbf{Entity Expansion}}\;
	$\mathbb{E} \gets $ entityExpansion(refC)\;
	return $\mathbb{E}$\;
\end{algorithm}

\subsection{Skip-gram Features Extraction and Clustering} 
\label{model_clustering}
We preprocess the raw corpus and extract skip-gram features of seed words as \cite{Shen2017SetExpanCS} and \cite{rong2016egoset} do. Here skip-gram features are a sequence of words surrounding the seed word. Based on the distributional hypothesis \cite{mikolov2013distributed}, the semantics of a word is reflected by its neighboring skip-grams. We can derive different facets of a seed word by separating its skip-grams into different semantic clusters. 

Embedding is commonly used in NLP applications to represent rich semantic information of words and phrases. We obtain the embedding for each skip-gram by simply averaging the embedding of its component words. The derivation of skip-gram embedding is another interesting research question, but it is not our focus in this work. 

Now we cluster these skip-gram embeddings to discover different semantic facets of a seed word. Most clustering algorithms require the number of clusters as input, which deviates from our problem setting. Also, we note that the embedding usually lies in a high-dimension space (typically of dimension 100-300), which leads to the poor and unstable performance of most existing non-parametric clustering algorithms \cite{steinbach2004challenges} (\eg, MeanShift \cite{comaniciu2002mean}).

To solve the two main issues of instability and hard coded cluster numbers as mentioned above, we propose to tackle the high-dimensional embedding clustering problem by the affinity propagation algorithm \cite{frey2007clustering}. 
Specifically, we construct a complete weighted graph where each node represents a skip-gram, and the edge weight between each pair of nodes indicates the cosine similarity of their corresponding skip-gram embeddings. 
After the weighted graph of skip-grams is constructed, the affinity propagation algorithm \cite{frey2007clustering} is applied to find the best skip-gram clusters.\footnote{We set the preference to be -60.}

Empirical results demonstrate that the affinity propagation based skip-gram clustering is able to identify a reasonable number of semantic facets (c.f. Section~\ref{sec:numberoffacets}).
We think one possible reason is that affinity propagation takes a similarity graph as its input, 
while most other non-parametric clustering algorithms (\eg, MeanShift \cite{comaniciu2002mean}) take skip-gram embeddings as its input. 
In such a clustering task, we are only interested in semantic similarities between skip-grams and do not care about the complete information of the skip-grams (\eg, semantic and syntactic information). 
Therefore, though skip-gram embeddings contain more information than a similarity graph, the information serves more as ``noise" and less as useful information. 
Moreover, the robustness of affinity propagation is immune to the dimension of the embeddings, while others can be very sensitive to it. 
In our experiment, we find MeanShift is highly unstable if the dimension is greater than 30. 
Hence affinity propagation, which takes similarities between pairs of data points as input, serves for our needs well.

\subsection{Discovering Coherent Semantic Facets of A Seed Set} 
\label{model_merging}
After obtaining multiple skip-gram clusters for each seed, we then need to determine the coherent semantic facets among all seeds and generate the coherent skip-gram clusters. Take two seed words ``apple'' with facets \emph{fruit} and \emph{company}, and ``orange'' with facets \emph{fruit} and \emph{color} as an example, their coherent semantic facet is \emph{fruit}.

The key is to determine whether a facet of seed word $A$ matches any facet of word $B$. 
Suppose that $A$ has $r$ skip-gram clusters $S_{A}=\{S_{A}^{(1)}, \ldots, S_{A}^{(r)}\}$, where cluster $S_{A}^{(i)}$ contains a set of skip-grams relevant to the $i$-th facet of $A$. 
Similarly, $B$ has $t$ skip-gram clusters $S_{B}=\{S_{B}^{(1)}, \ldots, S_{B}^{(t)}\}$. 
If $A$ and $B$ share $k$ facets, and they have $k$ pairs of matching clusters $\{(S_{A}^{(i_{1})}, S_{B}^{(j_{1})}), \ldots, (S_{A}^{(i_{k})}, S_{B}^{(j_{k})})\}$ accordingly. 
Therefore, these $k$ facets are jointly represented by these clusters: 

$S_{A,B}=\{S_{A}^{(i_{1})}\bigcup S_{B}^{(j_{1})}, \ldots,  S_{A}^{(i_{k})}\bigcup S_{B}^{(j_{k})}\}$.

We first measure the pairwise correlation of their skip-gram clusters (c.f.~Sec \ref{get_correlation}), and then make a matching decision on a pair of clusters (c.f.~Sec \ref{check_same_cluster}).

\subsubsection{Calculating correlation between two skip-gram clusters} \label{get_correlation}

Suppose that facet $A_{1}$ (one facet of word $A$) corresponds to a skip-gram cluster $\mathbf{X}=[\mathbf{x}_{1}; \ldots; \mathbf{x}_{m}]$ with $m$ skip-gram vectors, where $\mathbf{x}_{i}\in\mathbb{R}^{d}$. Similarly, facet $B_{1}$ (one facet of word $B$) corresponds to a cluster $\mathbf{Y}=[\mathbf{y}_{1}; \ldots; \mathbf{y}_{n}]$ with $n$ skip-gram vectors, where $\mathbf{y}_{j}\in\mathbb{R}^{d}$. Two clusters $\mathbf{X}$ and $\mathbf{Y}$ are from different seed words, and we want to measure their correlation in order to decide whether they correspond to the same semantic facet.

To measure their correlation, we find the semantic sense which $\mathbf{X}$ and $\mathbf{Y}$ have in common. Inspired by the idea of compositional semantics \cite{hermann2014multilingual,socher2014grounded}, we set the sense vector to the linear combination of skip-gram vectors. 

Suppose that the sense vector $\mathbf{u}$ from cluster $\mathbf{X}$ and the sense vector $\mathbf{v}$ from $\mathbf{Y}$ are the sense shared by the two clusters. Therefore, the common sense vectors should be highly correlated, \ie, we want to find $\mathbf{u}$ and $\mathbf{v}$ so that their correlation is maximized. We formulate the following optimization problem (\ref{eq:skipgram}).
\begin{align}
	\label{eq:skipgram}
	\nonumber
	&\max\limits_{a,b} \frac{\mathbf{u}^{T}\mathbf{v}}{\lVert\mathbf{u}\rVert\cdot \lVert\mathbf{v}\rVert} \\
	\nonumber
	\text{s.t.}\quad &\mathbf{u} = \mathbf{Xa}, \\
	&\mathbf{v} = \mathbf{Yb},
\end{align}
where $\mathbf{a}\in \mathbb{R}^{m}$ and $\mathbf{b}\in\mathbb{R}^{n}$ are coefficient vectors.

Solving the problem (\ref{eq:skipgram}) by CCA \cite{hardoon2004canonical}, we can find their common sense vectors $\mathbf{u}^{*}$ and $\mathbf{v}^{*}$.
The semantic correlation $corr(\mathbf{X},\mathbf{Y})$ between cluster $\mathbf{X}$ and $\mathbf{Y}$ is defined as the correlation between these two sense vectors:
\begin{align}
	corr(\mathbf{X},\mathbf{Y}) = {\mathbf{u}^{*}}^{T}\mathbf{v}^{*}
\end{align}

\subsubsection{Matching facets of all seeds} \label{check_same_cluster}

After quantifying correlation for two skip-gram clusters, we cast it as a binary decision whether the cluster $\mathbf{X}$ of facet $A_{1}$ matches semantically with any facet of word $B$. 

We note that it is not a good way to decide the matching clusters by setting a hard correlation threshold, since the the numerical correlation range is word-specific.
It is easy to see that if a facet of seed $A$ (\eg, $A_{1}$) is of the same semantic class with a facet of seed $B$ (\eg, $B_{2}$), then $corr(A_{1}, B_{2})$ is higher than the correlation between $A_{1}$ and any other facets of seed $B$. Otherwise, the correlation of $A_{1}$ and all facets of seed $B$ should be equally small. 

Based on the intuition above, we define a relevance score below: 
\begin{align}
	rele(A_{1}, B) &= D_{KL}(\mathbf{Corr}(A_{1}, B), U)
\end{align}
where $U$ is uniform distribution, $D_{KL}$ is KL-divergence~\cite{kullback1951information}, and $\mathbf{Corr}(A_{1}, B) = softmax((corr(A_{1}, B_{1}), ..., corr(A_{1}, B_{m})))$, 

We then make the matching decision based on the relevance score $rele(A_1, B)$. The threshold of the relevance score is set to 0.25 empirically. Once the matching decision is satisfied, we find the best matching facet $B^*$ in word $B$ and generate one coherent skip-gram cluster $A_1 \bigcup B^*$. 
Finally, we do facet matching for all facets of word $A$ and obtain the resulting skip-gram clusters as coherent skip-gram clusters.

\noindent \textbf{Remarks}: If there are more than two seed words, we first discover coherent skip-gram clusters of two seeds and then use their coherent skip-gram clusters to match with the third seed and so on.

\subsection{Entity Expansion} 
\label{model_expansion}
The coherent skip-gram clusters of different facets are used to expand the seed set. 
Traditionally, researchers expands entites from a group of skip-gram clusters based on graph-based approaches \cite{Wang2007LanguageIndependentSE}, entity matching approaches \cite{Shen2017SetExpanCS}, distributional hypothesis \cite{rong2016egoset} and iterative approaches \cite{wang2008iterative,He2011SEISASE}. 
As a distinction, we tackle the problem by a Masked Language Model (MLM), which leverages the discriminative power of the BERT model and contextual representations \cite{devlin2019bert}.

For each skip-gram denoted as $sg$, we compute the MLM probability $h_{c,sg}$ for each word candidate $c$ in the vocabulary by a pre-trained BERT model.\footnote{We use the `bert-base-uncased' pre-trained model from \url{https://huggingface.co/transformers/model\_doc/bert.html\#bertformaskedlm}.} 
Therefore, given a set of skip-grams, the weight $w_{c}$ of a word candidate $c$ is calculated as: 
$w_c = \sum_{sg'}h_{c, sg'}$. 
The final entity expansion process simply ranks all word candidates by their weights.

\section{Experiments}
\label{sec:exp}

\begin{table*}[t!]
	\centering
	\scriptsize
	\caption{Evaluation using MMAP (``recall"), PMAP (``precision") and BMAP (``F1 score").}
		\begin{tabular}{ccccccccccc}
			\toprule
			& &\multicolumn{3}{c}{\textbf{MMAP@\textit{l}}} & \multicolumn{3}{c}{\textbf{PMAP@\textit{l}}} & \multicolumn{3}{c}{\textbf{BMAP@\textit{l}}}\\ 
			\cmidrule{3-11} 
			& & \textbf{$l$=5} & \textbf{$l$=10} & \textbf{$l$=20} \space\space\space\space& \textbf{$l$=5} & \textbf{$l$=10} & \textbf{$l$=20} \space\space\space\space& \textbf{$l$=5} & \textbf{$l$=10} & \textbf{$l$=20}   \\
			\midrule
			\multicolumn{1}{c|}{\textbf{Single-}}& \textbf{word2vec} & 0.323 & 0.283 & 0.252    \space\space\space\space   & 0.552 & 0.499 & 0.448 \space\space\space\space& 0.390 & 0.352 & 0.316\\
			\cmidrule{2-11} 
			\multicolumn{1}{c|}{\textbf{Faceted}} & \textbf{SEISA} &  0.345 & 0.301 & 0.268 \space\space\space\space& 0.550 & 0.503 & 0.455 \space\space\space\space& 0.408 & 0.368 & 0.331  \\
			\cmidrule{2-11} 
			\multicolumn{1}{c|}{}& \textbf{SetExpan} & 0.373 & 0.337 & 0.304 \space\space\space\space& 0.605 & 0.563 & \textbf{0.512} \space\space\space\space& 0.448 & 0.413 & 0.374  \\
			\midrule
			\multicolumn{1}{c|}{\textbf{Multi-}} & \textbf{Sensegram} & 0.312 & 0.301 & 0.275 \space\space\space\space& 0.479 & 0.443 & 0.398\space\space\space\space & 0.359 & 0.343 & 0.314  \\
			\cmidrule{2-11} 
			\multicolumn{1}{c|}{\textbf{Faceted}} & \textbf{EgoSet} & 0.446 & 0.390 & 0.325 \space\space\space\space& 0.306 & 0.261 & 0.206 \space\space\space\space& 0.335 & 0.292 & 0.236 \\
			\cmidrule{2-11} 
			\multicolumn{1}{c|}{}& \textbf{\our} & \textbf{0.477} & \textbf{0.414} & \textbf{0.366} \space\space\space\space& \textbf{0.643} & \textbf{0.573} & 0.507 \space\space\space\space& \textbf{0.531} & \textbf{0.469} & \textbf{0.414}\\
			\midrule
			\multicolumn{1}{c|}{\textbf{Ablations}} & \textbf{\our-k ($k$=2)} & 0.420 & 0.364 & 0.326 \space\space\space\space& 0.607 & 0.540 & 0.494 \space\space\space\space& 0.478 & 0.422 & 0.383 \\ 
			\cmidrule{2-11} 
			\multicolumn{1}{c|}{} & \textbf{\our-k ($k$=3)} & 0.454 & 0.406 & 0.360 \space\space\space\space& 0.624 & 0.562 & 0.505 \space\space\space\space& 0.504 & 0.455 & 0.407 \\ 
			\bottomrule
		\end{tabular}
	\label{table:all_results}
\end{table*}

Our model targets the corpus-based entity set expansion problem, and thus we evaluate its performance on a local corpus. 

\noindent \textbf{Dataset}: We evaluate our approach, \our, on the dataset in \cite{rong2016egoset}. 
The dataset contains 56 million articles (1.2 billion words) retrieved from English Wikipedia 2014 Dump and 150 human-labeled multi-faceted queries. 

\subsection{Evaluation Metric}
It is considerably complicated to properly evaluate multi-faceted set expansion task due to different number of facets between the generated result and the ground truth. 
Previous work \cite{rong2016egoset} adopted the following Mean Mean Average Precision (MMAP) measure: 
$$ MMAP@\textit{l} = \frac{1}{M_q} \sum_{m=1}^{M_q} AP_l(B_{qi^*}, G_{qm}),$$
where 
$M_{q}$ is the number of facets for query $q$ in the ground truth; 
$G_{qm}$ is the ground truth set of $m$-th facet for $q$; 
$B_{qi^*}$ is the output facet that best matches $G_{qm}$, 
and $AP_l(c,r)$ represents the average precision of top $l$ entities in a ranked list $c$ given an unordered ground truth set $r$.
This metric measures the coverage of ground truth sets by the generated sets.

However, it does not penalize additional noisy facets in generated sets and thus it is biased towards the model that generates more facets. For example, a model generating 30 facets with 3 relevant facets achieves higher MMAP than another model generating 3 facets with 2 relevant facets. One can ``cheat" the performance by generating as many facets as possible.

To overcome the intrinsic limitation of MMAP, we, inspired by \cite{goldberg2010measuring,chinchor1992muc}, propose a new metric, Best-Matching Average Precision (BMAP) to capture both the purity of generated facets and their coverage of ground truth facets. 
Our metric is defined as follows: 
$$BMAP@\textit{l} =  HMean(MMAP@\textit{l}, PMAP@\textit{l}),$$
$$PMAP@\textit{l} = \frac{1}{F_q} \sum_{f=1}^{F_q} AP_l(B_{qf}; G_{qi^*}),$$
where 
$F_q$ is the number of facets in generated output; 
$B_{qf}$ is the $f$-th output ranked list for query $q$; 
$G_{qi^*}$ is the ground truth facet that best matches $B_{qf}$. Here $HMean(a,b) = \frac{2ab}{a+b}$ is the harmonic mean of $a$ and $b$.

Our proposed BMAP metric not only evaluates how well generated facets match the ground truth by $MMAP@\textit{l}$ but also penalizes low-quality facets by $PMAP@\textit{l}$. Intuitively, $MMAP@\textit{l}$ measures ``recall" to capture how many ground truth results has been discovered, while $PMAP@\textit{l}$ measures ``precision" to capture the fraction of good facets in the generated output. Accordingly, $BMAP@\textit{l}$ measures ``F1 score" to leverage ``precision" and ``recall". 
Results are reported by averaging all queries for each dataset.

\subsection{Baselines}

The following approaches are evaluated: 

\begin{itemize}
	\item \textbf{word2vec\footnote{\url{https://code.google.com/p/word2vec}}} \cite{Mikolov2013DistributedRO}: We use the ``skip-gram'' model in word2vec to learn the embedding vector for each entity, and then return $k$ nearest neighbors of the seed words.
	
	\item \textbf{SEISA} \cite{He2011SEISASE}: An entity set expansion algorithm based on iterative similarity aggregation. It uses the occurrence of entities in web list and query log as entity features. In our experiments, we replace the web list and query log with skip-gram features.
	
	\item \textbf{SetExpan\footnote{\url{https://github.com/mickeystroller/SetExpan}}} \cite{Shen2017SetExpanCS}: A corpus-based set expansion that selects quality context features for entity-entity similarity calculation and expand the entity sets using rank ensemble. 
	
	\item \textbf{EgoSet} \cite{rong2016egoset}: The only existing work for multi-faceted set expansion. It expands word entities from skip-gram features, and then clusters the expanded entities into multiple sets by the Louvain community detection algorithm. 
	
	\item \textbf{Sensegram\footnote{\url{https://github.com/uhh-lt/sensegram}}} \cite{pelevina-EtAl:2016:RepL4NLP}: 
	We learn different embeddings for each word's different senses and return $k$ nearest neighbors for each embedding. 
	
	\item \textbf{\our-k}: A variant of \our which replaces Affinity Propagation with $k$-means clustering algorithm for skip-gram clustering. 
\end{itemize}

\subsection{Results}
\label{sec:quanresults}

We compare the performance of \our against all baselines using MMAP (``recall"), PMAP (``precision") and BMAP (``F1 score"). 
As shown in Table~\ref{table:all_results}, \our achieves the highest scores in most cases and outperforms all other baselines in BMAP and MMAP.

It is worth mentioning that EgoSet achieves decent results in MMAP. However, it generates too many noisy facets, which deteriorate PMAP and the overall performance BMAP. We will further discuss this phenomenon in Sec.~\ref{sec:numberoffacets}. 

It is also interesting to note that single-faceted baselines (\ie, SetExpan, SEISA) have much stronger PMAP performance than multi-faceted baselines. This is because by generating a single cluster of the most confident expansion results, they usually match with one ground truth cluster very well and thus achieve high PMAP (``precision") value. 

In the ablation analysis, it is worth noting that \our, even without pre-determined number of clusters, performs better than \our-k. We experiment the number of clusters $k$ to be 2 and 3, which are the mode and the mean of the number of clusters of the ground truth respectively. 
We think the poor performance is because forcing skip-grams into a fixed number of clusters will induce clustering noise. 
Furthermore, the noise will propagate and be enlarged in the skip-gram cluster fusion step and the entity expansion step, and therefore leads to bad performance. 


\subsection{Number of Facets Identified}
\label{sec:numberoffacets}

We explore the number of facets identified by different multi-faceted set expansion methods. Specifically, we adopt $l_1$ and $l_2$ distances.

{
	\small
	\begin{eqnarray*}
		\mbox{$l_1$ distance} = \sum_{q \in Q} |\mbox{GT}_q - \mbox{Gen}_q| \\
		\mbox{$l_2$ distance} = \sqrt{\sum_{q \in Q} (\mbox{GT}_q - \mbox{Gen}_q)^2}
	\end{eqnarray*}
}

Here $Q$ is all queries, $\mbox{GT}_q$ and $\mbox{Gen}_q$ are the number of facets that ground truth has and the number of facets that the corresponding model identifies for query $q$, respectively. 

The distance measurement are summarized in Table \ref{table:2}. \our generates closer number of facets to the ground truth, compared to EgoSet, demonstrating about 65\% reduction of the $l_1$ distances and 45\% reduction of the $l_2$ distances. 

\begin{table}[h!]
	\centering
	\small
	\caption{Distance between the number of facets identified and the number of facets the ground truth has.}
	\begin{tabular}{ccc}
		\toprule
		& $l_1$ distance \space\space&\space\space $l_2$ distance \\
		\midrule
		EgoSet & 783 & 78.02  \\
		\midrule
		\our   & \textbf{277}  & \textbf{43.05} \\
		\bottomrule
	\end{tabular}
	\label{table:2}
\end{table}

\begin{table*}[t]
	\centering
	\caption{Case study on comparison between \our and EgoSet.}
	\includegraphics[width=0.99\linewidth]{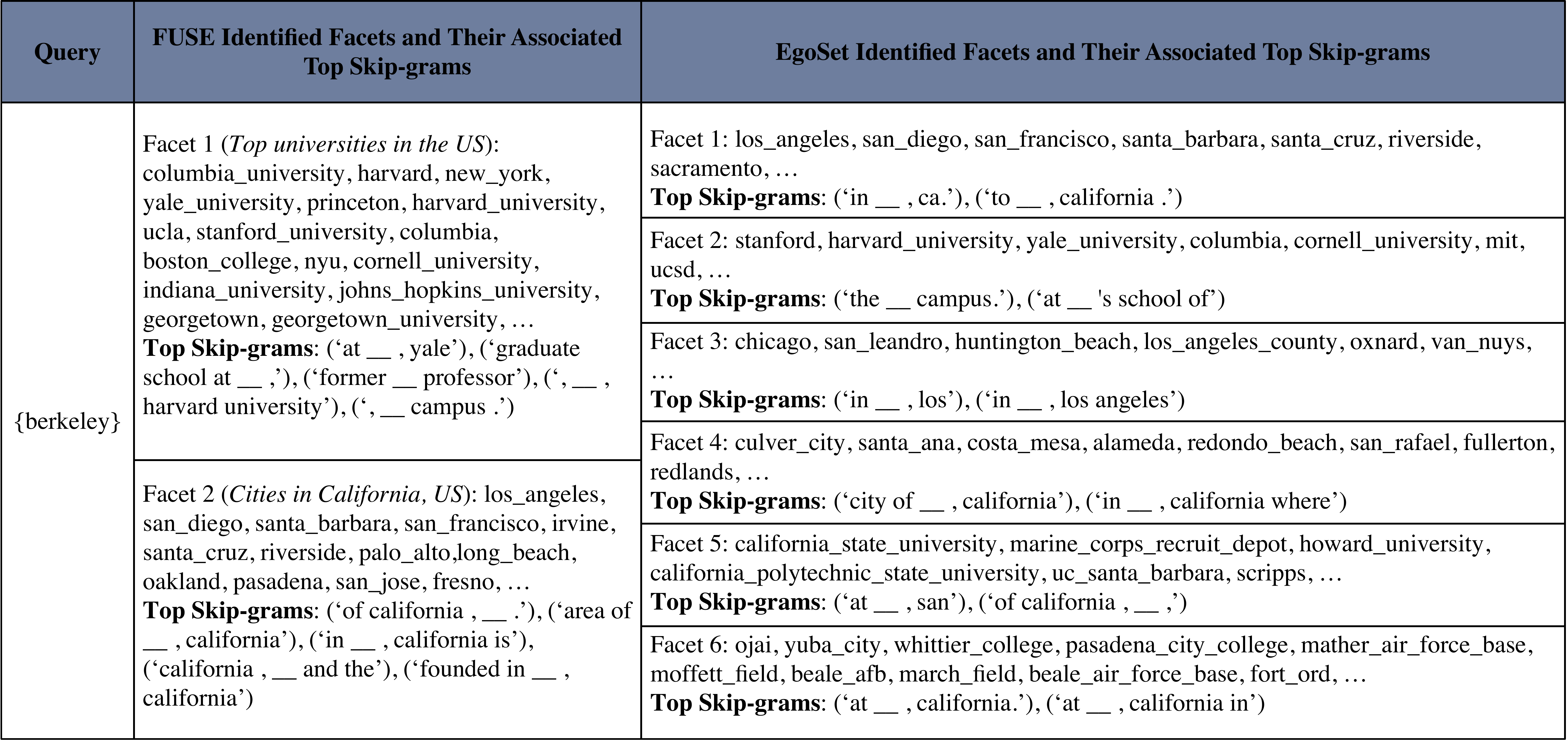}
	\label{table:Egoset-FUSE}
\end{table*}

To further explore identified facets between \our and EgoSet, we present one case study of query \{``berkeley"\} in Table \ref{table:Egoset-FUSE}. 
\our generates two facets and each facet has its distinctive semantic meaning (one for \textit{Top universities in the US} and one for \textit{Cities in California, US}), while EgoSet generates too many scattered facets, with less distinctiveness between facets. 
One of the reasons is that EgoSet performs clustering on expanded entities while \our performs clustering on skip-grams. 
Skip-grams, consisting of multiple words, are usually of more clear semantics compared to entities themselves (\eg, the entity ``columbia" can be either \textit{Universities}, \textit{Cities} or \textit{Rivers}, while the skip-gram ``graduate school at \_\_" is more clear to be \textit{Universities}.). 
Therefore, clustering on skip-grams is an easier process and results in better performance.

\subsection{Case Studies: Multi-Faceted Setting}
\label{sec:casestudy}

Table~\ref{fig:4} shows intermediate results of \our by listing top skip-grams of each semantic facet. 
It is worth noting that even the ground truth may not present a full coverage of semantic facets of given seeds. 
For example, as shown in Case~4, the ground truth only includes semantic facet \textit{Animals}. 
Our system also finds another meaningful semantic facet \textit{Tributaries}.\footnote{Beaver River, Elk River and Bear River are tributaries of Pennsylvania, Mississippi River and the Great Salt Lake, respectively.}
The query \{``Chongqing"\} shown in Case~5 is another example, where the ground truth again fails to capture the semantic facet of \textit{War-related Major Cities}.\footnote{Chongqing was the second capital of Chinese nationalist party during the war.}

\begin{table*}[ht]
	\centering
	\caption{Case studies on top skip-grams for each semantic facet. The concept name of each facet is in bold.}
	\includegraphics[width=0.99\linewidth]{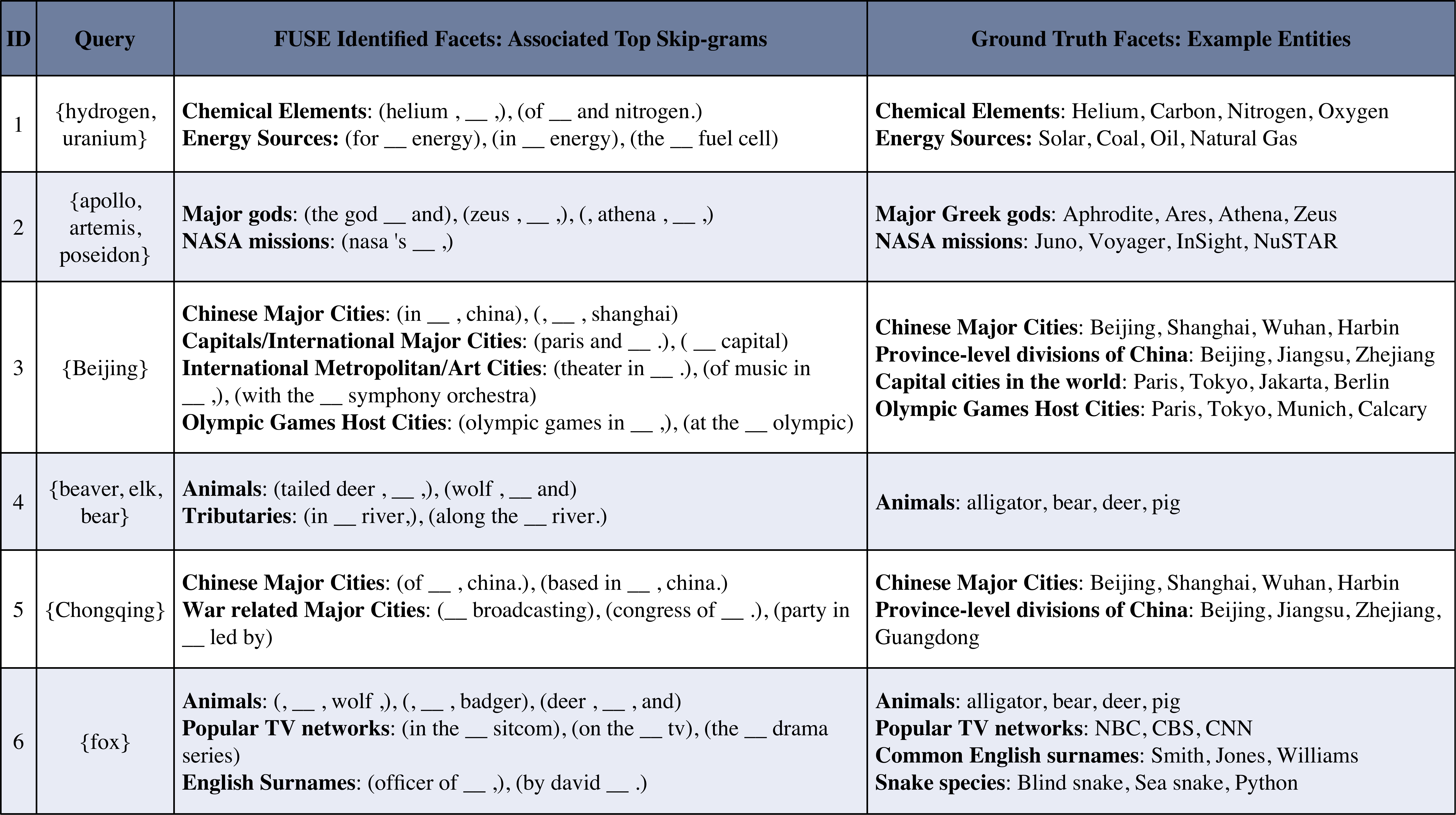}
	\label{fig:4}
\end{table*}

\section{Single-Faceted Set Expansion}
\label{sec:discussion}

In previous sections, we have demonstrated that \our has favorable performance against state-of-the-art systems in expanding multiple semantic facets of a seed set. 
Yet, it's more common in real-life cases that a seed set has one single semantic facet (especially when more seeds are provided). 
In this subsection, we inspect the performance of \our on single-faceted cases. 

In the singled-faceted set expansion task, there is exactly one semantic facet from a seed set.
However, if one or more words in the seed set are ambiguous, such ambiguity will introduce entities related to noisy facets, and thus hurt the quality of the expanded set. 
For example, the seed set \{``apple'', ``amazon''\} has only one semantic facet corresponding to \textit{Technology Companies}, however, the seed ``apple'' is an ambiguous word and has a noisy facet, \ie,  \textit{Fruits}. 
Most existing systems~\cite{tong2008system,Wang2007LanguageIndependentSE,wang2008iterative,Pantel2009WebScaleDS,sarmento2007more,He2011SEISASE,Wang2015ConceptEU,chen2016long,Shen2017SetExpanCS,Mamou2018TermSE,rong2016egoset} first extract all contextual features (\eg, skip-grams) from the entire seed set and then rank the keyword candidates based on the contextual features. 
To the best of our knowledge, none of them denoises contextual features from the noisy facet (\ie, \textit{Fruits}). 
Therefore, they are likely to generate entities related to the facet of \textit{Fruits} (\eg, ``pear", ``banana"), despite the fact that the seed set \{``apple", ``amazon"\} has only one semantic facet. 

In contrast, \our is robust to such lexical ambiguity, since we discover the shared coherent semantic facet across all seeds and expand entities by relevant contextual features only. 
From this example, one can clearly see that even for single-faceted set expansion, it is also critical to resolve the lexical ambiguity and identify the common facet among seeds.

To gain deeper insights in the single-faceted setting, we present a case study on the seed set \{``apple'', ``amazon''\} in Table~\ref{fig:general}. We highlight those noisy entities resulting from seed ambiguity (\ie, the \textit{Fruits} sense) in bold, bright red. 
As expected, EgoSet has 2 facets that contain entites from the \textit{Fruits} sense and SetExpan suffers from the noisy facet issue too. 
As it has shown, \our performs favorably against previous approaches in single-faceted set expansion too, in that it is robust to semantic ambiguity by extracting the coherent semantic facet shared by all seed words. 

\begin{table*}
	\centering
	\caption{Case studies on single-faceted set expansion: \{``apple", ``amazon"\}.}
	\includegraphics[width=0.99\linewidth]{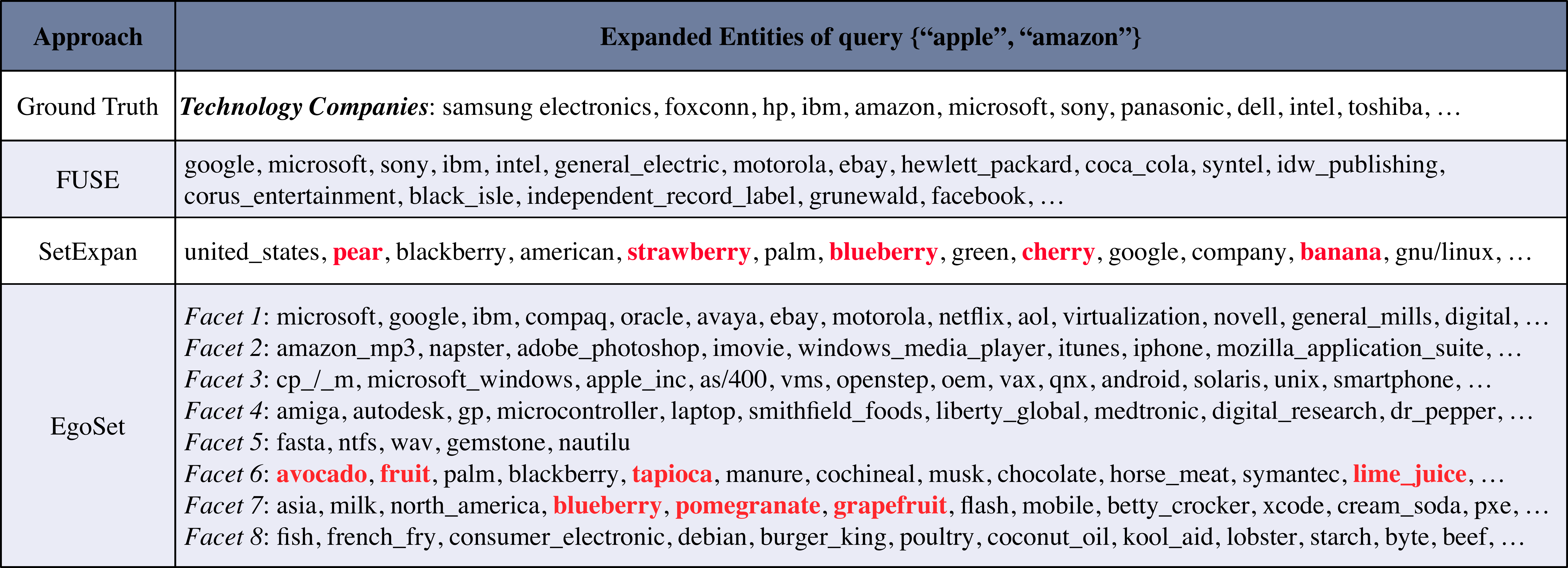}
	\label{fig:general}
\end{table*}

\section{Related Work}
Early work on entity set expansion, including \emph{Google Sets} \cite{tong2008system}, \emph{SEAL} \cite{Wang2007LanguageIndependentSE}, and \emph{Lyretrail} \cite{chen2016long} 
submits a query consisting of seed entities to a general-domain search engine (\eg, Google) and then mines the returned, top-ranked web pages. 
These approaches depend on the external search engine and require costly online data extraction. 

Later studies shift to the \emph{corpus-based} set expansion setting, where sets are expanded within a given domain-specific corpus. 
For example, \cite{Pantel2009WebScaleDS} compute the semantic similarity between two entities based on their local contexts and treat the nearest neighbors around the seed entities as the expanded set.
\cite{He2011SEISASE} further extend this idea by proposing an iterative similarity aggregation function to calculate entity-entity similarity using query logs and web lists besides free text. 
More recently, \cite{Shen2017SetExpanCS,Shen2018HiExpanTT} propose to compute semantic similarity using only selected high-quality context features, and \cite{Mamou2018SetExpanderET,Mamou2018TermSE} develop SetExpander system to leverage multi-context term embedding for entity set expansion. 
All the above attempts, however, assume the input seed entities belong to one unique, clear semantic class, and thus largely suffer from the multi-faceted nature of these seeds -- they could represent multiple semantic meanings.

To resolve the ambiguity of seeds, \cite{Wang2015ConceptEU} propose to utilize the target semantic facet name and then retrieve its most relevant web tables. However, it requires the exact name of the target semantic facet and outputs the one semantic facet of entities only. This does not accomplish multi-faceted set expansion in nature. 

The only attempt towards multi-faceted set expansion is EgoSet~\cite{rong2016egoset}, to the best of our knowledge. 
EgoSet first extracts quality skip-gram features to construct an entity-level ego-network, and then perform the Louvain community detection algorithm on the ego-network to extract entities for different facets. 
Finally, it combines them with external knowledge (\ie, user-generated ontologies) to generate final output. 
Although \our may appear to be similar to EgoSet at first glance, we highlight key differences and significance below: 

\begin{itemize}
	\item \textbf{Key challenges of multi-faceted set expansion}: We identify the key challenge of multi-faceted set expansion to be discovery of shared semantic facets from a seed set. While in EgoSet, noisy facets that are relevant only to partial seeds are also generated. As a result, it fails to solve multi-seed single-faceted cases (\eg, \{``apple", ``amazon"\}) and multi-seed multi-faceted cases (\eg, Fig.~\ref{fig:1}). 
	
	\item \textbf{External knowledge}: EgoSet requires user-created ontology (obtained from Wikipedia) as external knowledge. While these semi-structured web tables and ontologies are helpful for disambiguation, they are not always available for domain-specific corpus. \our relies on free text only and thus, can be applied in a more general setting. 
	
	\item \textbf{Clustering over skip-grams}: EgoSet adopts clustering (community detection) over expanded entities, while \our adopts clustering over skip-grams. 
	Clustering over entities usually leads to mediocre results in non-parametric settings, since any expanded entity can be ambiguous. 
	However, skip-grams, consisting of multiple words, are usually of more clear semantics and much easier to be clustered compared to entities themselves (demonstrated in Sec.~\ref{sec:numberoffacets}).
	In additional, EgoSet adopts hard clustering on entities, which ignores the nature that the same entity may fall into different facets (\eg, ``Paris" should appear in both sets of \textit{National Capitals} and \textit{Olympic Games Host Cities} in Fig.~\ref{fig:1}), while the design of \our naturally allows the same entities to appear in multiple facets. 
\end{itemize}

In a more general way, our work is also related to word sense disambiguation~\cite{taghipour2015semi,raganato2017neural,raganato2017word,iacobacci2016embeddings,pelevina-EtAl:2016:RepL4NLP}. The major difference is that our work aims to find the coherent semantic facets of all seed words and achieve entity expansion from the coherent skip-gram clusters.

\section{Conclusion}
We identify the key challenge of the problem -- \textit{multi-faceted set expansion} and have proposed a novel and effective approach, \our, to address it. 
By extracting and clustering skip-grams for each seed, identifying coherent semantic facets of all seeds, and expanding entity sets for each semantic facet, 
\our is capable of identifying coherent semantic facets, generating quality entity set for each facet, and therefore outperforms previous state-of-the-art approaches significantly.  

The proposed framework \our is general in that it achieves quality set expansion in both multi-faceted and single-faceted settings. 
In particular, it, for the first time, solves the case where different seeds have different multi-facetedness. 
For future work, we plan to explore other skip-gram clustering approaches and coherent semantic facet discovery algorithms.

\subsubsection{Acknowledgments}
Research was sponsored in part by US DARPA KAIROS Program No. FA8750-19-2-1004 and SocialSim Program No.  W911NF-17-C-0099, National Science Foundation IIS 16-18481, IIS 17-04532, and IIS-17-41317, and DTRA HDTRA11810026. 

\bibliographystyle{splncs04}

\end{document}